\documentclass{article}

\usepackage[final]{neurips_2024}

\usepackage[utf8]{inputenc} %
\usepackage[T1]{fontenc}    %
\usepackage{hyperref}       %
\usepackage{url}            %
\usepackage{booktabs}       %
\usepackage{amsfonts}       %
\usepackage{nicefrac}       %
\usepackage{microtype}      %
\usepackage{xcolor}         %
\usepackage{amsmath}
\usepackage{soul}
\usepackage{multirow}
\usepackage{graphicx}
\usepackage{bm}
\usepackage{amsmath}
\usepackage{graphicx}
\usepackage{multirow}
\usepackage{enumerate}
\usepackage{marvosym}

\title{On the Identifiability of Hybrid Deep Generative Models: Meta-Learning as a Solution}

\author{
  Yubo Ye \thanks{Work was done as a visiting exchange student at Rochester Institute of Technology.}  \\
  Zhejiang University \\
  \texttt{22230131@zju.edu.cn}
  \And
  Maryam Toloubidokhti \\
  Rochester Institute of Technology \\
  \texttt{mt6129@rit.edu}
  \And
  Sumeet Vadhavkar \\
  Rochester Institute of Technology \\
  \texttt{sv6234@rit.edu}
  \And
  Xiajun Jiang \\
  Rochester Institute of Technology \\
  \texttt{xj7056@rit.edu}
  \And
  Huafeng Liu \Letter \\
  Zhejiang University \\
  \texttt{liuhf@zju.edu.cn}
  \And
  Linwei Wang \\
  Rochester Institute of Technology \\
  \texttt{linwei.wang@rit.edu} \\
}

\begin{document}

\maketitle
\begin{abstract}
 The interest in leveraging physics-based inductive bias in deep learning has resulted in recent development of \textit{hybrid deep generative models (hybrid-DGMs)} that integrate known physics-based mathematical expressions in neural generative models. 
To identify these hybrid-DGMs requires inferring parameters of the physics-based component along with their neural component. 
 The identifiability of these hybrid-DGMs, however, has not yet been theoretically probed or established. How does the existing theory of the un-identifiability of general DGMs apply to hybrid-DGMs? 
 What may be an effective approach to construct a hybrid-DGM with theoretically-proven identifiability?
This paper provides the first theoretical probe into the identifiability of hybrid-DGMs, and presents meta-learning as a novel solution to construct identifiable hybrid-DGMs. 
On synthetic and real-data benchmarks, we provide strong empirical evidence for 
the un-identifiability of existing hybrid-DGMs using unconditional priors, and strong identifiability results of the presented meta-formulations of hybrid-DGMs.
Code is available at \url{https://github.com/GoForIt-YYB/Meta-Hybrid-VAE}.
\end{abstract}

\section{Introduction}

There has been increasing interest in integrating mathematical expressions of known physics with neural-network functions for hybrid, or gray-box, modeling \cite{APHYNITY, physcis-integrated-VAE, Physics-Augmented-Learning,rubust-hybrid-vae,Qian21}. 
The incorporation of physics-based inductive bias has the potential to improve the generalizability and interpretability of deep learning \cite{rubust-hybrid-vae}, 
while the expressive and flexible neural functions offer the opportunity to fill the gap in 
our prior knowledge of physics (or \textit{expert models}) \cite{rubust-hybrid-vae,Qian21}. 
Recent progress in particular has started to see principled developments of hybrid deep generative models (DGM) where the data-generation process is described by a hybridization of physics-based and neural components. 
Both of these two components are then identified from data via, for instance, amortized variational inference \cite{physcis-integrated-VAE,rubust-hybrid-vae}. 
These hybrid-DGMs have shown benefits of interpretability, generalizability, and 
out-of-distribution (OoD) robustness, 
especially when learning to model physics systems from observational data \cite{physcis-integrated-VAE,rubust-hybrid-vae}.
An important premise of these successes 
is that 
the hybrid expression of the physical systems can be accurately identified from the observed data: 
for example, 
if we can identify the parameters and/or physics governing observed pendulum movements, we could then use the identified generative models/parameters for predicting the future trajectory of the pendulum \cite{physcis-integrated-VAE}.

However, substantial research based on the theory of nonlinear independent component analysis (nonlinear ICA)
have established that 
a naive DGM or latent variable model is un-identifiable  \cite{iVAE}. 
Active research has also reported 
various ways to construct an identifiable latent variable model, 
primarily via constructing a conditionally independent generative model 
\cite{TCL, PCL, GCL,iVAE}. 
Most of the progress has been established for the identifiability of latent processes underlying time-series data, leveraging the temporal dependency of the latent variables \cite{PCL} or observed domain index for non-stationary segments \cite{TCL} 
to construct the conditional generative model. 
For static latent variables, 
the construction of conditionally independent generative model 
has mainly relied on the introduction of auxiliary observations such as class or domain labels  \cite{GCL}. 

How does the hybrid expression of a DGM impact its (un)-identifiability? 
Furthermore, 
since hybrid-DGMs are often intended  
to
model physics systems with  unknown parameters to be identified from data, 
what may be an effective approach 
to construct a hybrid-DGM with 
theoretically-proven identifiability? 
Unfortunately, 
despite the increasing interest 
in hybrid-DGMs, 
their identifiability remains unexplored. 
This identifiability however is critical for the intended use of a hybrid DGM, \textit{e.g.,} to predict beyond the observations used to identify the model, or to be robust in out-of-distribution (OOD) generalizations. 

Recent efforts in hybrid-DGMs indeed have increasingly noted the challenges in properly learning the two components within the hybrid models, especially in ensuring a non-trivial solution to the physics-based component due to the presence of a highly-expressive neural component \cite{APHYNITY,physcis-integrated-VAE}. 
Various strategies have been presented, such as regularizing the expressiveness of the neural component \cite{physcis-integrated-VAE}, trajectory-based and adaptive optimization to allow automatic complexity adjustment for the neural component \cite{APHYNITY}, and expert-augmentation leveraging the physics-based component to improve generalization to distributional shifts \cite{rubust-hybrid-vae}. 
No existing works, 
however, have probed or established the identifiability of hybrid-DGMs, theoretically or empirically. 

In this work, we present the first theoretical probe of the identifiability of hybrid-DGMs. 
Importantly, forgoing the need of \textit{observed} auxiliary variable to condition the generative model, 
we present meta-learning as a novel solution to establish the identifiability of hybrid-DGMs. 
To this end, we present a learn-to-identify formulation for meta-learning hybrid-DGMs, 
and theoretically establish the identifiability of these meta-hybrid-DGMs via conditional independence of the DGM given 
few-shot context samples. %
We further provide empirical identifiability results of these meta-hybrid-DGMs in a variety of synthetic and real datasets generated from physics systems, in contrast to existing non-meta hybrid-DGMs where identifiability cannot be established without observed auxiliary variables. 
Finally, we note that while this work focuses on hybrid-DGMs, 
the presented meta-learning formulation provides a general new condition to construct identifiable DGMs beyond the hybrid setting.

\section{Related Works}

\paragraph{Hybrid-DGMs:} A number of hybrid models have emerged in various domains to 
combine physics-based functions with neural networks, to compensate for unknown components in the known physics. 
While earlier approaches \cite{hwangbo2019learning, zeng2020tossingbot, golemo2018sim} focused on using neural networks to learn the residual between the measurements and those obtained from physics-based simulation, tighter integrations have been presented by, \textit{e.g.,} NeuralSim \cite{NeuralSim} 
and universal differential equations (UDEs) \cite{UDE}, to 
include neural networks as different components within a physics-based function. 
Most of these earlier works are focused on hybrid modeling in specific application domains. 

More recently, several hybrid-DGMs have been presented \cite{physcis-integrated-VAE,APHYNITY,rubust-hybrid-vae,Physics-Augmented-Learning}, 
with a focus on 
properly learning the physics component within the hybrid-DGM so it is not overpowered by the expressive neural component. 
In physics-integrated hybrid VAE \cite{physcis-integrated-VAE}, 
this was achieved by a regularized learning method 
that controls the expressiveness of the neural component while preserving the semantics of the physics-based component. 
Similarly, regularization of the expressiveness of the neural component is considered in \cite{APHYNITY}. 
In \cite{rubust-hybrid-vae}, expert augmentation is proposed to improve the OOD robustness of these hybrid-DGMs by fine-tuning on synthetic OOD samples, generated by sampling outside the training distribution of the physics-based component within the trained hybrid model. 

Despite these efforts in better learning the two separate components within hybrid-DGMs,  
the identifiability of hybrid-DGMs has not been theoretically probed nor established. 
Empirically, no existing works have evaluated these hybrid-DGMs from the lens of identifiability metrics. 
Theoretically, 
no existing works have examined the 
conditions for constructing 
identifiable hybrid-DGMs.
This work will take the first step to bridge these gaps, 
establish the un-identifiability of 
naive unconditional hybrid-DGMs, 
and present meta-learning as a novel solution to 
construct identifiable hybrid-DGMs.

\paragraph{Nonlinear ICA and Identifiability of DGMs:} 
With nonlinear ICA, strong identifiability results have been established for unsupervised latent variable models \cite{TCL,PCL,GCL,iVAE,SlowVAE,NCTRL,SNICA,HMNICA}. 
Fundamentally, construction of identifiable generative models 
is achieved by defining independence structure when conditioned on observed auxiliary variables \cite{GCL}. 
Much progress has been made in identifying latent dynamic processes generating observed time series, including leveraging conditional independence of the DGM given time segment index \cite{TCL}, %
temporal dependence \cite{PCL}, 
sparse temporal encoding \cite{SlowVAE}, 
and even unobserved inferred domain labels for non-stationary time series \cite{NCTRL,SNICA,HMNICA}. 
For static latent variables, identifiability is mainly established via the introduction of auxiliary observations such as class or domain labels \cite{GCL}. 
In specific, iVAE \cite{iVAE} established the identifiability of DGMs 
assuming an exponential families conditional distributions given observed auxiliary variables. 

The reliance on observed conditioning variable, however, limits the applicability
of these identifiability results to hybrid-DGMs: as hybrid-DGMs are intended for modeling physics system with unknown and often continuous parameters, 
it is not clear what \textit{observed} auxiliary label may be available to construct an identifiable DGM. 
This work will forego this assumption of observed auxiliary variable and present meta-learning as a novel condition to construct identifiable DGMs.  
While presented for hybrid-DGMs, 
this connection between meta-learning and the identifiability of DGMs 
is a novel contribution to general non-hybrid DGMs that has not been constructed in the existing literature.

\section{Problem Formulation: Unidentifiability of Hybrid-DGMs} 
\label{sec:problem}

Let $\mathbf{x}\in \mathbb{R}^n$ be the observed random variables and $\mathbf{z} \in \mathbb{R}^d$ be the latent variables generating $\mathbf{x}$, with the following generation process: 
\begin{equation}
\label{eqn:generation}
    p(\mathbf{x}, \mathbf{z})=p(\mathbf{z})p(\mathbf{x}|\mathbf{z}), \quad \quad 
        p(\mathbf{x})=\int_{\mathbf{z}}p(\mathbf{x} ,\mathbf{z})d\mathbf{z} 
\end{equation}
where $p(\mathbf{z})=\prod_{i=1}^dp(z_i)$ represents the prior distribution 
of independent generative factors $z_i$, 
$p(\mathbf{x}|\mathbf{z})$ is the likelihood function which can be defined based on the mixing function $\mathbf{x} = \mathcal{F}(\mathbf{z})$. 

\paragraph{Construction of Hybrid-DGMs:}
When constructing a generative model for Equation \eqref{eqn:generation}, 
if we assume a known mathematical expression explaining the mechanism of the mixing function as $f_\text{P}$,
we will have a physics-based expression of this generation process, with unknown parameters of $f_\text{P}$ as 
physics-based latent variables $\mathbf{z}_\text{P}$.   
If we have no prior knowledge, 
we can describe the mixing function 
with a neural network $f_\text{N}$ 
with abstract representation of the latent variables as $\mathbf{z}_\text{N}$ 
-- the latter is the foundation of many successful DGMs including variational autoencoders (VAEs) \cite{kingma2013auto}. 

When understanding complex systems in many domains, 
recent years have seen an increasing interest in the middle-ground of the above two scenarios:
some prior knowledge in the form of a mathematical expression $f_\text{P}$ of physics exists about the data observed, 
yet often inexact with unknown gaps to the actual data-generating mechanisms. 
This motivated recent developments of hybrid-DGMs where 
the mixing function $\mathbf{x} = \mathcal{F}(\mathbf{z})$ is described as a combination of a physics-based and neural component, $f_\text{P}$ and $f_\text{N}$, respectively, 
with corresponding physics-based 
and abstract latent variables $\mathbf{z}_\text{P}$ and $\mathbf{z}_\text{N}$. 

As a concrete example, 
suppose that we observe the 
time-series of the angular position $\varphi$ of a damped forced pendulum system, 
$\mathbf{x}=[\varphi_0, \varphi_1,\dots,\varphi_T]$ where $\varphi_i=\varphi(i\Delta t)$, 
governed by: %
\begin{equation}
    \label{Eq1}
    \frac{d^2\varphi(t)}{dt^2}
    +\omega^2\sin\varphi(t)+\xi \frac{d\varphi(t)}{dt}-A\cos(2\pi\phi t)=0
\end{equation}
where the first two terms describe the physics of an \textit{ideal} pendulum system, 
the third term the damping effect, 
and the last term the external force. 
Parameters 
$\omega, \xi, A, \text{ and }\phi$ specify different systems governed by the same physics, 
representing the independent latent generative factors. 

Suppose that we choose to leverage our prior knowledge about the \textit{ideal} pendulum physics 
$f_\text{P}(\mathbf{x}; \mathbf{z}_\text{P}) 
=\ddot{\varphi}+\mathbf{z}_\text{P}^2 \sin \varphi$
with unknown parameter $\omega$.
To bridge its potential gap with actual data-generating physics, we then choose to complete it with a neural-network expression of an ODE
$f_{\text{N}_\theta}(\mathbf{x};  \mathbf{z}_\text{N})$, 
parameterized by $\theta$ and with an abstract latent variable $\mathbf{z}_\text{N}$: 
while abstract, 
we hope $f_{\text{N}_\theta}$ 
to absorb the effect of the last two terms in 
the data-generating physics 
(Equation \eqref{Eq1}
with $\mathbf{z}_\text{N}$ representing its remaining data-generating parameters $\xi, A, \text{ and }\phi$.  
This gives rise to a hybrid mixing function $\mathbf{x} = \mathcal{F}[f_\text{P}, f_{\text{N}_\theta}; \mathbf{z}_\text{P}, \mathbf{z}_\text{N}] =$ 
 $\text{ODEsolve}[f_\text{P}(\mathbf{x};\mathbf{z}_\text{P})+f_{\text{N}_\theta}(\mathbf{x};\mathbf{z}_\text{N})=0]$. 
 
 Assuming Gaussian observation noises, 
 we obtain the likelihood of the Hybrid-DGMs as:
\begin{equation}
    \label{Eq:unconditional prior}
p_\theta(\mathbf{x}|\mathbf{z}_\text{P}, \mathbf{z}_\text{N})=\mathcal{N}(\mathbf{x}|\mathcal{F}
    [ f_\text{P}, f_{\text{N}_\theta}; \mathbf{z}_\text{P}, \mathbf{z}_\text{N} ] , \Sigma_x)
\end{equation}
with prior distribution for the latent variables assumed to be Gaussian as:
\begin{equation}
\label{eqn:zprior}
    p(\mathbf{z}_\text{P}):=\mathcal{N}(\mathbf{z}_\text{P}|\mu_\text{P}, \sigma^2_\text{P}\mathbf{I}), \quad  \quad
    p(\mathbf{z}_\text{N}):=\mathcal{N}(\mathbf{z}_\text{N}|0,\mathbf{I})
\end{equation}
where the mean 
$\mu_\text{P}$ and 
variance $\sigma^2_\text{P}$ 
for $p(\mathbf{z}_\text{P})$ can be 
defined by prior knowledge due to its physical meanings. 
Note that $\mathbf{z}_\text{P}$ will be directly interpretable and physically meaningful as they will be semantically grounded to the parameters of the physics
model $f_\text{P}$; 
$\mathbf{z}_\text{N}$, in comparison, 
will be abstract 
but need to absorb effect of the varying parameters 
underlying the missing physics
$f_\text{N}$.

The data-generating process of the hybrid-DGM with parameter $\theta$ can thus be defined as: 
\begin{equation}
\label{eqn:generation:hybrid}
    p_\theta(\mathbf{x}, \mathbf{z}_\text{P}, \mathbf{z}_\text{N})=p_\theta(\mathbf{x}|\mathbf{z}_\text{P}, \mathbf{z}_\text{N})p(\mathbf{z}_\text{P})p(\mathbf{z}_\text{N}), \quad \quad  p_\theta(\mathbf{x})=\int \int p_\theta(\mathbf{x}, \mathbf{z}_\text{P}, \mathbf{z}_\text{N})d\mathbf{z}_\text{P} d\mathbf{z}_\text{N}
\end{equation}

While described in the context of a concrete example, 
Equations \eqref{Eq:unconditional prior}-\eqref{eqn:generation:hybrid} represent a general expression of hybrid-DGMs where the hybrid-mixing function 
$\mathbf{x} %
=\mathcal{F}[f_\text{P}, f_{\text{N}_\theta}; \mathbf{z}_\text{P}; \mathbf{z}_\text{N}]$ 
can be designed in various forms 
(and not limited to the hybrid-ODE described in the example above).

\paragraph{Unidentifiability of Hybrid-DGMs:}

Now assume we have access to 
data $\mathcal{D}=\left \{\mathbf{x}^{(1)}, \dots,  \mathbf{x}^{(N)} \right \}$ generated by $p_{\theta^*}(\mathbf{x}, \mathbf{z}_\text{P}, \mathbf{z}_\text{N})$, 
where $\theta^*$ is the true but unknown data-generating parameter. 
For an identifiable DGM, 
our goal
is to learn 
$\theta$ such that 
the likelihood 
$p_\theta^*(\mathbf{x}|\mathbf{z}_\text{P}, \mathbf{z}_\text{N})$, 
the priors 
$p(\mathbf{z}_\text{P})$ and $p(\mathbf{z}_\text{N})$, 
and the posteriors 
$p_\theta^*(\mathbf{z}_\text{P}|\mathbf{x})$ 
and $p_\theta^*(\mathbf{z}_\text{N} |\mathbf{x})$ are all correctly recovered.  

In practice, because we only observe $\mathbf{x}$ without access to the latent variables 
$\mathbf{z}_\text{P}$ or $\mathbf{z}_\text{N}$, 
we optimize $\theta$ to 
match the marginal density of $\mathbf{x}$, \textit{e.g.,} 
via amortized variational inference to maximize the evidence lower bound (ELBO) of $p_{\theta}(\mathbf{x})$ in VAEs, 
such that: 
\begin{equation}
    p_{\theta}(\mathbf{x}) \approx p_{\theta^*}(\mathbf{x})
\end{equation}

Unfortunately, even if 
the above optimization is done perfectly (\textit{i.e.,} 
$p_{\theta}(\mathbf{x}) = p_{\theta^*}(\mathbf{x})$
), 
there is no guarantee that 
$p_\theta^*(\mathbf{x}|\mathbf{z}_\text{P}, \mathbf{z}_\text{N})$, 
$p_\theta^*(\mathbf{z}_\text{N} |\mathbf{x})$, 
or
$p_{\theta^*}(\mathbf{x}, \mathbf{z}_\text{P}, \mathbf{z}_\text{N})$ are correctly identified. 
As shown in \cite{iVAE},
for these densities to be recovered based on matching 
$p_{\theta}(\mathbf{x})$, 
the DGM needs to satisfy:
\begin{equation}
    \label{Eq3}
    \forall (\theta, \Tilde{\theta}): p_{\theta}(\mathbf{x})=p_{\Tilde{\theta}}(\mathbf{x}) \Rightarrow \theta = \Tilde{\theta}
\end{equation}

In another word, the matching of $p_{\theta}^*(\mathbf{x})$ needs to imply a unique solution of ${\theta}^*$ to be obtained. 
In reality, however, 
it has been shown in \cite{iVAE} that DGMs with unconditional prior are unidentifiable. 
We refer the readers to the important theorems %
and the proof in \cite{iVAE,GCL, Exponential-Families} for more details, 
but to what extent is the hybrid-DGM affected by this theory of identifiability?

At a concept level, 
let us first 
consider the 
neural component ($\mathbf{z}_\text{N}$
and $f_{\text{N}_\theta}$) within the 
hybrid-DGM. 
For $\mathbf{z}_\text{N}$ of any distributions, there are always nonlinear transformations that change its values but not its distributions \cite{iVAE}. 
In a highly expressive neural network, 
this transformation can be learned 
in the likelihood 
$p_\theta(\mathbf{x}|\mathbf{z}_\text{N})$. 
This will give the same fit of $p_\theta(\mathbf{x})$, but 
non-unique solutions of $p_\theta(\mathbf{x}|\mathbf{z}_\text{N})$ and $p_\theta(\mathbf{z}_\text{N}|\mathbf{x})$, 
resulting in non-identifiable neural component within the hybrid-DGM.

For the physics-based component ($\mathbf{z}_\text{P}$
and $f_{\text{P}_\theta}$),
this theory does not hold
because the pre-specified form of $f_{\text{p}}$ can not trivially accommodate any transformations applied to $\mathbf{z}_{\text{P}}$ without 
affecting the fitting of observations. 
In practice, however, 
because both components jointly contribute to 
$p_\theta(\mathbf{x}|\mathbf{z}_\text{P}, \mathbf{z}_\text{N})$ as defined in Equation \eqref{Eq:unconditional prior}, 
the optimization of $\mathbf{z}_\text{P}$
and $f_{\text{P}_\theta}$ maybe overpowered by the overly-expressive neural component, 
rendering trivial solutions to $\mathbf{z}_\text{P}$
and $f_{\text{P}_\theta}$.

In other words,
identification of hybrid-DGMs is inflicted by two fundamental challenges: 
1) the theoretical un-identifiability of its neural component, 
and 
2) this un-identifiable but highly expressive component overpowering the 
identification of the otherwise identifiable physics-based component. 
While existing works 
\cite{physcis-integrated-VAE,APHYNITY,Physics-Augmented-Learning} 
have focused on the second challenge, 
we address the problem from its fundamentals 
at the un-identifiability 
of $\mathbf{z}_\text{N}$
and $f_{\text{N}_\theta}$: 
we introduce a novel condition to construct 
identifiable hybrid-DGMs,  
and establish their identifiability both theoretically and empirically.

\section{Learn-to-Identify Hybrid-DGMs via Meta-Learning}

Based on nonlinear ICA theory \cite{GCL}, 
an identifiable DGM can be constructed by 
conditionally 
independent 
latent variables $p(\mathbf{z}|\mathbf{u})$: in the literature, 
the conditioning variable $\mathbf{u}$ is often assumed to be additionally-observed side information  
such as class labels or domain indexes \cite{iVAE}. 
Here, we show that meta-learning formulations offer 
an opportunity to construct an identifiable DGM, 
with $\mathbf{u}$ taking forms of 
$k$-shot 
context samples sharing the same 
data-generation process as the query samples. %

\paragraph{Construction of Identifiable Meta-Hybrid-DGMs:} 
Given is a set of $k$ context samples, $\mathcal{D}^s$, 
sharing the same 
data-generation process as query samples $\mathbf{x}$, 
we define the generation of $\mathbf{x}$ as:
\begin{equation}
\label{eqn:generation:meta-hybrid}
    p_\phi(\mathbf{x}, \mathbf{z}_\text{P}, \mathbf{z}_\text{N}|\mathcal{D}^s)=p_\theta(\mathbf{x}|\mathbf{z}_\text{P}, \mathbf{z}_\text{N})p(\mathbf{z}_\text{P})p_\zeta(\mathbf{z}_\text{N}|\mathcal{D}^s), \quad  p_\phi(\mathbf{x}|\mathcal{D}^s)=\int \int p_\phi(\mathbf{x}, \mathbf{z}_\text{P}, \mathbf{z}_\text{N}|\mathcal{D}^s)d\mathbf{z}_\text{P} d\mathbf{z}_\text{N}
\end{equation}
where $\phi = \{ \theta, \zeta\}$, 
$p_\theta(\mathbf{x}|\mathbf{z}_\text{P}, \mathbf{z}_\text{N})$ is defined in Equation \eqref{Eq:unconditional prior}, 
and
the conditional prior $p_\zeta(\mathbf{z}_N|\mathcal{D}^s)$ %
is assumed to be factorized exponential family distributions \cite{exp-distribution}:
\begin{equation}
    \label{Eq:conditioned prior}
    p_\zeta(\mathbf{z}_N|\mathcal{D}^s)=p_{\mathbf{T},\pmb{\lambda}_\zeta}(\mathbf{z}_\text{N}|\mathcal{D}^s)=\prod_{i=1}^{d} \frac{Q_i(z_{\text{N},i})}{Z_{i}(\mathcal{D}^s)}\exp \left [ 
    \mathbf{T}_i(z_{\text{N},i})^\top \pmb{\lambda}_{\zeta, i}(\mathcal{D}^s)
    \right ]
\end{equation}
where $Q_i$ is the base measure, $Z_i(\mathcal{D}^s)$ is the normalizing constant, $\mathbf{T}_i=( T_{i,1}, \dots, T_{i,e} )$ are the sufficient statistics 
that are fixed (not estimated). $\pmb{\lambda}_i^\zeta(\mathcal{D}^s)=(\lambda_{i,1}^\zeta(\mathcal{D}^s),\dots, \lambda_{i,e}^\zeta(\mathcal{D}^s))$ are the corresponding parameters conditioning on $\mathcal{D}^s$.
In this paper, we realize this conditioning through a composite of a neural network 
$h_\zeta(\mathbf{x}^s)$, parameterized by $\zeta$, for individual context samples $\mathbf{x}^s\in \mathcal{D}^s$, and an
averaging function across all context samples:  
\begin{equation}
    \label{eqn:meta_end}   \bm{\lambda}_\zeta(\mathcal{D}_s)=\frac{1}{\left | \mathcal{D}^s \right |} \sum_{\mathbf{x}^s\in \mathcal{D}^s}h_\zeta(\mathbf{x}^s)
\end{equation} 

Note that exponential families have universal approximation capabilities, thus this commonly-used assumption is not very restrictive \cite{Exponential-Families}.

\paragraph{Amortized Variational Inference:}
To enable inference over the hybrid-DGM in Equation \eqref{eqn:generation:meta-hybrid}, 
we approximate the posterior density 
$p(\mathbf{z}_\text{P}|\mathbf{x})$ as $q_\eta(\mathbf{z}_\text{P}|\mathbf{x})$
and $p_\zeta(\mathbf{z}_\text{N}|\mathbf{x}, \mathcal{D}^s)$ as $q_\zeta(\mathbf{z}_\text{N}|\mathbf{x} \cup \mathcal{D}^s)$, 
the latter 
realized with
the same network as defined in \eqref{Eq:conditioned prior} 
but with the additional input $\mathbf{x}$ in addition to $\mathcal{D}^s$.

Formally, we cast the variational inference into a meta-learning formulation. Consider a dataset $\mathcal{D}$ with $M$ similar but distinct data-generation process: $\mathcal{D}=\left \{ \mathcal{D}_m \right \}_{m=1}^M$. For each $\mathcal{D}_m$, we consider disjoint context samples $\mathcal{D}_m^s=\left \{ \mathbf{x}^{s,1}, \mathbf{x}^{s,2},\dots, \mathbf{x}^{s,k} \right \}$ and query samples $\mathcal{D}_m^q=\left \{ \mathbf{x}^{q,1}, \mathbf{x}^{q,2},\dots, \mathbf{x}^{q,l} \right \}$, where $k \ll l$. Instead of maximizing the marginal likelihood of $\mathbf{x}$ for all $\mathbf{x} \in \mathcal{D}$, we formulate a meta-objective to learn to maximize the marginal likelihood $p(\mathbf{x}^q|\mathcal{D}_m^s)$ of all query samples $\mathbf{x}^q \in \mathcal{D}_m^q$ when conditioned on support set  $\mathcal{D}_m^s$, for all underlying data-generation processes $m \in \left \{ 1,2,\dots, M \right \}$:
\begin{equation}
    \label{Eq:meta-objective}
\begin{split}
\sum_{m=1}^M \sum_{\mathbf{x}^q\in\mathcal{D}_m^q}\log p(\mathbf{x}^q|\mathcal{D}^s_m) \ge
& \sum_{m=1}^M \sum_{\mathbf{x}^q\in\mathcal{D}_m^q} 
\{ \mathbb{E}_{q_\eta(\mathbf{z}_\text{P}|\mathbf{x}^q), q_\zeta(\mathbf{z}_\text{N}|\mathbf{x}^q \cup \mathcal{D}_m^s)}
[\log p_\theta(\mathbf{x}^q|\mathbf{z}_\text{P}, \mathbf{z}_\text{N})] \\
& -\text{KL}(q_\eta(\mathbf{z}_\text{P}|\mathbf{x}^q)||p(\mathbf{z}_\text{P}))
-\text{KL}( q_\zeta(\mathbf{z}_\text{N}|\mathbf{x}^q \cup \mathcal{D}_m^s) || p_\zeta(\mathbf{z}_\text{N}|\mathcal{D}_m^s) ) \}
\end{split}
\end{equation}
where the first term 
represents likelihood of the meta-hybrid-DGM on query sample $\mathbf{x}^q \in \mathcal{D}_m^q$, 
and the last two terms represent Kullback–Leibler divergences 
between the prior and posterior densities of $\mathbf{z}_\text{P}$ and  $\mathbf{z}_\text{N}$, respectively, 
where densities of $\mathbf{z}_\text{N}$ are conditioned on 
context samples $\mathcal{D}_m^s$.  
The maximization of Equation \eqref{Eq:meta-objective} is performed over $\{\eta, \zeta, \theta \}$ in an episodic training where, in each episode, 
the division of $\mathcal{D}_m^s$
and $\mathcal{D}_m^q$  is shuffled within each $\mathcal{D}_m$. 
The likelihood is calculated by the reparameterization trick \cite{kingma2013auto}, 
while the two KL-terms are calculated analytically. 

Note that in Equation \eqref{Eq:meta-objective},
the meta-hybrid-DGM is
formulated on the neural component within the hybrid-DGM
to establish its identifiability. 
While the identifiability of the physics-based component is not hinged on this construction, 
in practice, 
it is possible to extend the
conditional prior (Equation \eqref{Eq:conditioned prior}) 
to $\mathbf{z}_\text{P}$ to obtain $p_\zeta(\mathbf{z}_\text{P}, \mathbf{z}_\text{N}|\mathcal{D}_m^s)$, 
leading to an alternative meta-formulation:
\begin{equation}
    \label{Eq:meta-objective2}
\begin{split}
\sum_{m=1}^M \sum_{\mathbf{x}^q\in\mathcal{D}_m^q}\log p(\mathbf{x}^q|\mathcal{D}^s_m) \ge
& \sum_{m=1}^M \sum_{\mathbf{x}^q\in\mathcal{D}_m^q} 
\{ \mathbb{E}_{q_\zeta(\mathbf{z}_\text{P},\mathbf{z}_\text{N}|\mathbf{x}^q \cup \mathcal{D}_m^s)}
[\log p_\theta(\mathbf{x}^q|\mathbf{z}_\text{P}, \mathbf{z}_\text{N})] \\
&
-\text{KL}( q_\zeta(\mathbf{z}_\text{P}, \mathbf{z}_\text{N}|\mathbf{x}^q \cup \mathcal{D}_m^s) || p_\zeta(\mathbf{z}_\text{P}, \mathbf{z}_\text{N}|\mathcal{D}_m^s) ) \}
\end{split}
\end{equation}

We will empirically compare the two formulations above, and
demonstrate 
how the identifiability of $\mathbf{z}_\text{P}$ and $\mathbf{z}_\text{N}$ is each affected by the proposed meta-construction of the hybrid-DGMs.

\section{Identifiability Theory for Meta-Hybrid-DGMs} 

We consider the meta-formulation a novel condition to ensure the identifiability of unsupervised DGMs. Here, we show that, 
built on nonlinear ICA and the theory of identifiability for conditional-VAEs established in \cite{iVAE}, the presented meta-hybrid-DGMs are identifiable.

For readability, below we 
collect the formulations of the hybrid-DGM from previous sections:
\begin{equation}
    \label{Eq:generating}
    p_\phi(\mathbf{x}, \mathbf{z}_\text{P}, \mathbf{z}_\text{N}|\mathcal{D}^s)=p_\theta(\mathbf{x}|\mathbf{z}_\text{P}, \mathbf{z}_\text{N})
    p(\mathbf{z}_\text{P})
    p_{\mathbf{T},\pmb{\lambda}_\zeta}(\mathbf{z}_\text{N}|\mathcal{D}^s)
\end{equation}
\begin{equation}
    \label{Eq:mixing}    p_\theta(\mathbf{x}|\mathbf{z}_\text{P}, \mathbf{z}_\text{N}) =p_\varepsilon(\mathbf{x}-\mathcal{F}
    [ f_\text{P}, f_{\text{N}_\theta}; \mathbf{z}_\text{P}, \mathbf{z}_\text{N} ])
\end{equation}
\begin{equation}
    \label{Eq:exp-distribution}
   p_{\mathbf{T},\pmb{\lambda}_\zeta}(\mathbf{z}_\text{N}|\mathcal{D}^s)=\prod_{i=1}^{d} \frac{Q_i(z_{\text{N},i})}{Z_{i}(\mathcal{D}^s)}\exp \left [ 
    \mathbf{T}_i(z_{\text{N},i})^\top \pmb{\lambda}_{\zeta, i}(\mathcal{D}^s)
    \right ]
\end{equation}
where $\phi = (\theta, \mathbf{T}, \pmb{\lambda}_\zeta)$ are model parameters and $\pmb{\lambda}_\zeta$ is defined in Equation \eqref{eqn:meta_end}. 
Equation \eqref{Eq:mixing} 
gives a more general definition than Equation \eqref{Eq:unconditional prior}
where 
$\varepsilon$ is an independent noise variable. %

\paragraph{Definition 1: }\textit{Let $\sim$ be an equivalence relation on the parameter space $\Phi$. We say the DGM in \ref{eqn:generation} is $\sim$-identifiable if $p_{\phi}(\mathbf{x})=p_{\Tilde{\phi}}(\mathbf{x}) \Rightarrow \phi \sim \Tilde{\phi}$}

\paragraph{Definition 2: }\textit{Let $\sim$ be the equivalence relation on $\Phi$ defined as follows: }
\begin{equation}
    (\theta,\mathbf{T}, \pmb{\lambda}_\zeta) \sim (\Tilde{\theta}, \Tilde{\mathbf{T}}, \pmb{\lambda}_{\Tilde{\zeta}}) \Leftrightarrow \exists \mathbf{A},\mathbf{c}: \mathbf{T}(\mathcal{F}_{\theta}^{-1}(\mathbf{x}))=\mathbf{A}\Tilde{\mathbf{T}}(\mathcal{F}_{\Tilde{\theta}}^{-1}(\mathbf{x}))+\mathbf{c}, \forall\mathbf{x} \in \mathcal{X}
    \label{eq:identifiable_A}
\end{equation}
\textit{where $\mathbf{A}$ is a $de \times de$ matrix and $\mathbf{c}$ is a vector of dimension $de$, 
$d$ being the dimension of $\mathbf{z}_\text{N}$ and 
$e$ the dimensionality of the sufficient statistics $\mathbf{T}_i$.
If $\mathbf{A}$ is invertible, we denote this relation by $\sim_A$.} 

Definition 2 establishes a specific equivalence relation that allows to recover the sufficient statistics of the generative model up to a linear matrix multiplication.  

\paragraph{Theorem 1: }\textit{Assume we observe data sampled from 
$p_\theta(\mathbf{x}, \mathbf{z}_\text{P}, \mathbf{z}_\text{N}|\mathcal{D}^s)$ 
as defined in 
Equations \eqref{Eq:generating}-\eqref{Eq:exp-distribution}
with parameters 
$\phi = (\theta, \mathbf{T}, \pmb{\lambda}_\zeta)$.}
\textit{Assume the following holds:}
\textit{\begin{enumerate}[(i)]
    \item \textit{The set $\left \{ \mathbf{x} \in \mathcal{X} | \varphi_\varepsilon(\mathbf{x})=0 \right \}$ has measure zero, where $\varphi_\varepsilon$ is the characteristic function of the density $p_\varepsilon$ defined in 
$p_\varepsilon(\mathbf{x}-\mathcal{F}
    [ f_\text{P}, f_{\text{N}_\theta}; \mathbf{z}_\text{P}, \mathbf{z}_\text{N} ])$}.
    \item \textit{The hybrid mixing function $\mathcal{F}_\theta$ is injective.}
    \item \textit{The sufficient function $T_{i,j}$ are differentiable almost everywhere, and linearly independent on any subset of $\mathcal{X}$ of measure greater than zero.}
    \item \textit{There exist $de+1$ distinct context sets $\mathcal{D}^{s,0}, \mathcal{D}^{s,1},\dots, \mathcal{D}^{s,de}$ such that the $de \times de$ matrix $\mathbf{L}$ defined as follows is invertible:} %
\begin{equation}
    \mathbf{L} = (\pmb{\lambda}(\mathcal{D}^{s,1})-\pmb{\lambda}(\mathcal{D}^{s,0}), \dots, \pmb{\lambda}(\mathcal{D}^{s,de})-\pmb{\lambda}(\mathcal{D}^{s,0}))
\end{equation}
\end{enumerate}}

\textit{Then the parameters $\phi=(\theta, \mathbf{T}, \pmb{\lambda}_\zeta)$ are $\sim_A$-identifiable.} 

The proof for Theorem 1 directly builds on that presented in \cite{iVAE}.
Condition (iv) in Theorem 1 establishes that, 
if the generative factors in the neural component of the hybrid-DGM is 
$d$-dimensional each specified with $e$-dimensional sufficient statistics, 
we will need observations from a minimum of $de+1$ distinct and independent generation processes 
in order to identify the hybrid-DGM. 
Considering the pendulum-system example 
discussed in Section \ref{sec:problem} and Equation \eqref{Eq1}, 
since the neural component is generated by three independent system parameters 
$\xi, A, \text{ and }\phi$ ($d = 3$) and assuming Gaussian statistics ($e = 2$), 
the hybrid-DGMs will be identifiable if we have observations generated from 
more than $3\times 2 + 1 = 7$ distinct combination of parameter values for  
$\omega, \xi, A, \text{ and }\phi$ 
in the true data-generation process (Equation \eqref{Eq1}). We will empirically verify this condition in Section \ref{sec:exp}.

\section{Experiments and Results}
\label{sec:exp}

\paragraph{Data:} We considered three simulated 
and one real-world benchmarks for hybrid-DGMs, 
including three simulated physics systems
of forced damped pendulum \cite{physcis-integrated-VAE}, advection-diffusion system \cite{physcis-integrated-VAE} and double pendulum \cite{rubust-hybrid-vae}, 
and one real-world system double pendulum 
\cite{rubust-hybrid-vae}. 
For each of the three simulated physics systems, we randomly sampled the initial states and parameters of the governing function 
to generate observations. 
We refer to the governing functions used for the generation of data as \textit{full physics} functions, and design \textit{partial physics} functions for represent our imperfect prior knowledge about the observed data,  reflecting a variety of potentially additive and multiplicative errors -- these \textit{partial physics} functions are used as the $f_\text{P}$ in the hybrid models. 
In data generation, we varied the parameters in the components both present and absent in the prior physics. 
More details on each dataset can be found in 
Appendix \ref{app:data}.

\paragraph{Baselines:} 
We considered 
1) physics-integrated hybrid VAE \cite{physcis-integrated-VAE} %
and 
2) APHYNITY \cite{APHYNITY}, 
both using regularization to control the expressiveness of the neural component. 
Note that while the original formulation of APHYNITY does not have an inference network to identify $\mathbf{z}_\text{P}$ 
or $\mathbf{z}_\text{N}$ 
(making it intended for data generated from a single parameter setting), 
we added an inference network 
(that shares a similar architecture with the other two models) for fairness of comparison. 

\paragraph{Metrics:} With a focus on identifiability, 
we focus on the following metrics: 
1) for physics-based latent variable $\mathbf{z}_\text{P}$, 
we consider mean squared error (MSE) owing to its physical meaning; 
2) for abstract latent variable 
$\mathbf{z}_\text{N}$ modeled in the neural component, 
we consider the mean correlation coefficient (MCC) between the true data-generating parameters and 
$\mathbf{z}_\text{N}$ sampled from its learned posterior density. 
The calculation of MCC follows standard practice in the identifiability literature \cite{iVAE,GCL}: to evaluate $\sim_A$-identifiability, we calculate the weak MCC with the method described in \cite{ICE-BeeM}, 
where a high MCC provides evidence of successful identification. We also evaluate a more strong identifiable relation $\sim_P$ with strong MCC and the details can be found in  Appendix \ref{app:strongMCC}.

In addition to metrics on $\mathbf{z}_\text{P}$ and $\mathbf{z}_\text{N}$, 
we further introduce MSE metrics on the generated $\mathbf{x}$ in two distinct scenarios: 
1) \textit{reconstruction MSE} measures 
how well an observed $\mathbf{x}$ is reconstructed from the identified 
$\mathbf{z}_\text{P}$ and $\mathbf{z}_\text{N}$, 
and 
2) \textit{prediction MSE} measures how well the identified 
$\mathbf{z}_\text{P}$ and $\mathbf{z}_\text{N}$ can be used to 
generate outside the observed $\mathbf{x}$ used to identify them, 
\textit{e.g.,} for predicting over long time domains, or predicting for 
a different sample that 
comes from the same generation process but with different initial conditions. 
We use these two metrics to showcase that 
a good reconstruction does not guarantee a good identification, 
while the ability to use the identified latent variables to predict for different samples (other than observed) 
may be a  surrogate for identifiability. %

\begin{table}[]
\caption{Quantitative identifiability metrics for the presented meta-hybrid-VAE in comparison to physics-integrated hybrid-VAE \cite{physcis-integrated-VAE} and APHYNITY \cite{APHYNITY}, including MSE and MCC metrics on the latent variables, as well as MSE of generated $\mathbf{x}$ during reconstruction of observed samples (Rec) and prediction of unobserved samples (Pre).}
\resizebox{\linewidth}{!}{
\begin{tabular}{c ccc ccc}
\toprule
\multirow{2}{*}{} & \multicolumn{3}{c}{Forced Damped Pendulum}                     & \multicolumn{3}{c}{Advection-Diffusion System}\\
& \multicolumn{1}{c}{APHYNITY} & \multicolumn{1}{c}{Hybrid-VAE} & \multicolumn{1}{c}{Meta-Hybrid-VAE} & \multicolumn{1}{c}{APHYNITY} & \multicolumn{1}{c}{Hybrid-VAE} & \multicolumn{1}{c}{Meta-Hybrid-VAE} \\ \midrule
                                   
MSE of $\mathbf{z}_\text{P} \downarrow$       
&6.96(0.01)e-2   & 4.14(0.29)e-2  &  \bf{1.59(0.07)e-2}                                   
&1.9(0.3)e-2   & 7.12(5.32)e-4  &  \bf{1.34(0.75)e-4}                                    
\\ \midrule
MCC $\uparrow$              
&0.79(0.02)   & 0.59(0.03)  &  \bf{0.99(0.00)}                                    
&0.98(0.01)   & 0.94(0.00)  &  \bf{0.99(0.00)}    

\\ \midrule \midrule

MSE of $\mathbf{x}$ (Rec) $\downarrow$  
&5.2(0.01)e-2   &  \bf{2.66(0.03)e-2} &  2.91(0.06)e-2            
& 6.22(0.41)e-3   &  4.90(0.75)e-3 &  \bf{2.93(0.22)e-3}             
\\ \midrule
MSE of $\mathbf{x}$ (Pre)  $\downarrow$     
&2.37(0.67)   &  1.74(0.20)  &  \bf{6.85(1.11)e-2}                                   
& 8.25(0.58)e1  &  8.53(5.72)e-2 &  \bf{5.63(1.34)e-3} 
\\ \bottomrule
\end{tabular}}
\resizebox{\linewidth}{!}{
\begin{tabular}{c ccc ccc}
\toprule
\multirow{2}{*}{} & \multicolumn{3}{c}{Double Pendulum}                     & \multicolumn{3}{c}{Real Double Pendulum}\\
& \multicolumn{1}{c}{APHYNITY} & \multicolumn{1}{c}{Hybrid-VAE} & \multicolumn{1}{c}{Meta-Hybrid-VAE} & \multicolumn{1}{c}{APHYNITY} & \multicolumn{1}{c}{Hybrid-VAE} & \multicolumn{1}{c}{Meta-Hybrid-VAE} \\ \midrule

MSE of $\mathbf{z}_\text{P}$  $\downarrow$      
&4.58(0.04)e-1  & 3.97(0.06)e-1  &  \bf{3.85(0.19)e-1}                                    
& /  & /  & /                                     
\\ \midrule
MCC   $\uparrow$             
&0.50(0.00)   &  0.51(0.00) & \bf{0.98(0.00)}                                      
&  / & /  & /                                    \\ \midrule
\midrule
MSE of $\mathbf{x}$ (Rec)   $\downarrow$
&4.88(0.00)e-2   & 4.05(0.25)e-2  & \bf{3.83(0.06)e-2}             
&3.78(0.52)e-2   & \bf{2.20(0.14)e-3}  & 2.67(0.34)e-2            
\\ \midrule
MSE of $\mathbf{x}$ (Pre)  $\downarrow$     
&1.29(0.23)e1   & 5.52(0.06)  & \bf{2.80(0.48)e-1}                                     
&5.16(0.69)   & 1.87(0.13)  &  \bf{2.88(0.34)e-2}                                    
\\ \bottomrule
\end{tabular}}

\label{tab:allresults}
\end{table} 

\begin{figure}[t]
    \centering
    \includegraphics[width=\linewidth]{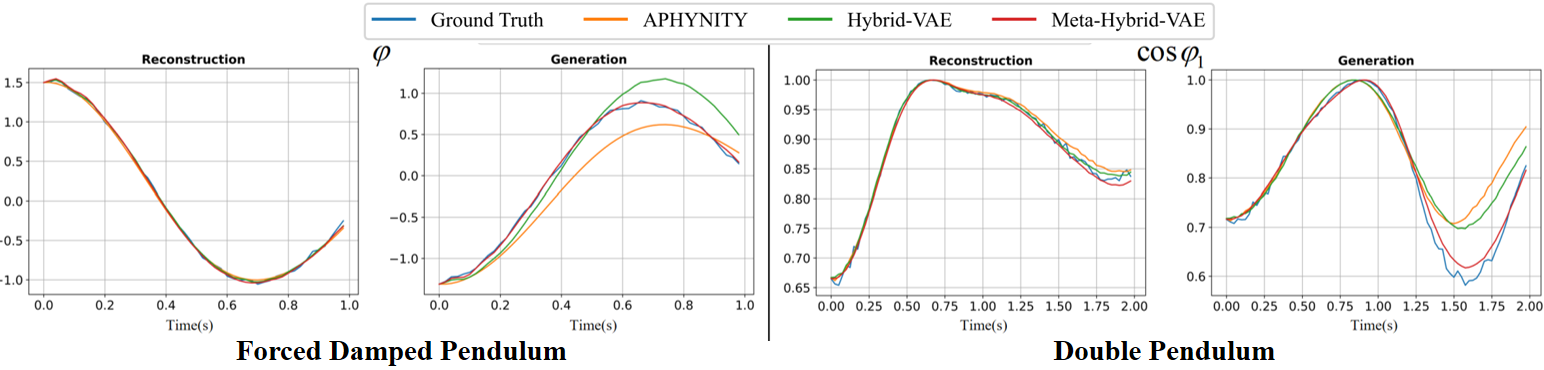}
    \caption{Visual results on synthetic data for reconstruction and prediction performances. }
    \label{fig:visual1}
\end{figure}
\subsection{Results on Synthetic Data} 
\label{subsec:synthetic}

\paragraph{Reconstruction \textit{vs}.\ Prediction Performance:}  
Table \ref{tab:allresults} summarizes 
quantitative identifiability metrics 
on the presented meta-hybrid-VAE 
in comparison to its baselines, 
where the prediction performance of all models 
is mentioned by their ability -- once identified -- to predict for a sample sharing the same data-generating parameters but different initial conditions. 
As shown, 
meta-hybrid-VAE significantly 
improved the identifiability of 
the abstract latent variable $\mathbf{z}_\text{N}$ (the highest MCC close to 1) and via which, 
moderately improved the accuracy of the 
physics-based latent variable 
$\mathbf{z}_\text{P}$ (the lowest MSE). 
Very importantly, 
while both baseline models identified 
$\mathbf{z}_\text{P}$ 
with a reasonable accuracy, 
their MCC values were significantly lower -- 
providing strong empirical support 
for the un-identifiability of the neural component of the hybrid-DGM as 
theorized in this paper. 

Also importantly, 
all hybrid models achieved comparable reconstruction MSE for observed $\mathbf{x}$'s, 
where the meta-hybrid-VAE did not necessarily produce the lowest MSE. 
The ability of the identified hybrid-DGM to predict for different samples other than observed, however, 
varied significantly: 
while the meta-hybrid-VAE was able to deliver a MSE comparable to the reconstruction task, 
the two baseline models saw a deterioration of MSE by two magnitudes. 
Fig.~\ref{fig:visual1} provides 
visual examples 
to demonstrate this performance difference between \textit{reconstruction} and \textit{prediction} tasks, 
stressing the importance to 
1) look beyond reconstruction performance for evaluating hybrid-DGMs, 
and 
2) consider prediction of unobserved samples 
as a potential surrogate for identifiability measures 
when ground-truth of latent variables is not available (real data settings in Section \ref{subsec:real}).

Interestingly, among the three datasets, 
the advection-diffusion system
appeared to be relatively simple to identify as the baselines did not exhibit as significant a degradation of performance in the more difficult 
identifiability metrics 
(MCC and prediction-MSE). 
This is potentially because 
the missing component from the prior physics has a small effect. %
In comparison, 
the MCC and prediction-MSE of the baseline models were significantly poorer in double pendulum, 
suggesting a difficult missing physics to be identified. 
Regardless, 
meta-hybrid-VAE was able to obtain significant margins of improvements compared to baselines across all datasets.

\begin{figure}[t]
    \centering
    \includegraphics[width=\linewidth]{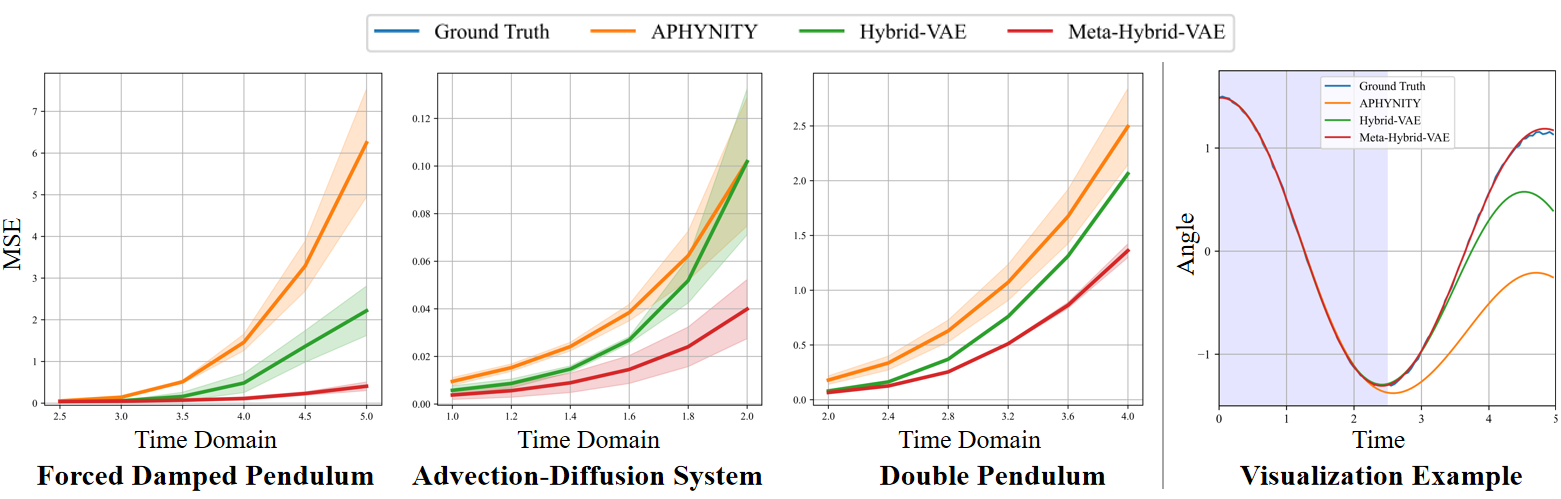}
    \caption{(Left) The performance of predictions over longer time domains; (Right) A visual example of forced damped pendulum system.The dark part is the training time domain, and the light part is the time domain outside training (longer time domain).}
    \label{fig:long_time}
\end{figure}

\paragraph{Predictions over Longer Time Domains:} 
Fig. \ref{fig:long_time} shows the prediction performance of the three models when predicting the trajectory of pendulum movement beyond the time domain used to identify the model, with a visual sample of forced damped pendulum system.
As shown, 
while all models' performance deteriorated as they predicted outside the time domain used in training, 
the presented meta-hybrid-VAE demonstrated significantly smaller drop in performance and slower rate of deterioration as the prediction horizon increased. This provides strong evidence regarding the importance of 
the identifiability, even for the neural component, of a hybrid model for the purpose of predictive tasks.

\begin{figure}[t]
    \centering
    \includegraphics[width=0.75\linewidth]{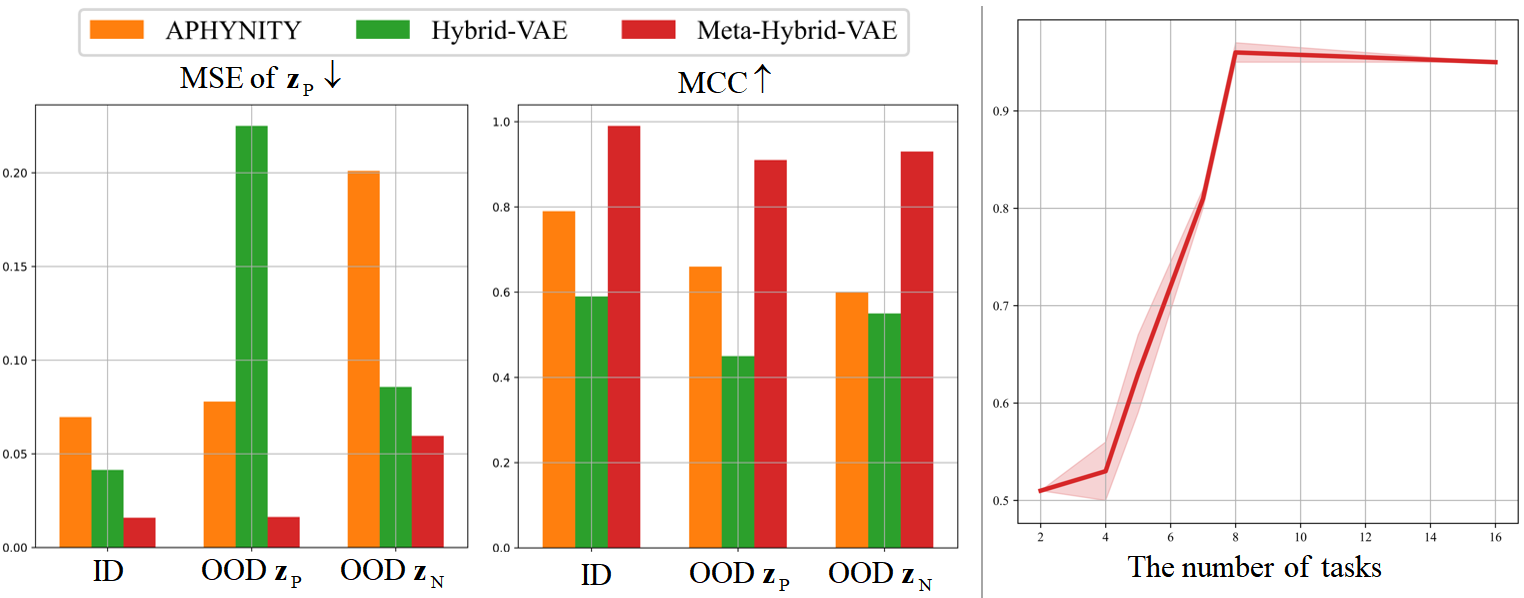}
    \caption{(left) Results of OOD; (right) Results of the condition for identifiability.}
    \label{fig:OOD}
\end{figure}

\paragraph{OOD Performance:} 
Figure \ref{fig:OOD} (left) summarizes the performance of the three models 
in settings where either the physics-based component $\mathbf{z}_\text{P}$ or the neural component $\mathbf{z}_\text{N}$ was outside the training distribution for forced damped pendulum system. 
Results demonstrated that, meta-hybrid-VAE obtained the strongest performance in OOD settings for both the identification of the physics and neural components. Details of the OOD parameters can be found in the Appendix \ref{subsec:p}.

\paragraph{Empirical Verification of the Condition for Identifiability:} 
Theorem 1-iv) stipulates that the identifiability of the hybrid-DGM depends on the number of 
independent conditional densities observed, in relation to the dimension of the latent factors to be identified. 
To experimentally verify this, 
we evaluated the identifiability metric MCC of the hybrid-DGM when learning over data generated from an increasing number of distinct parameter values governing the pendulum system (Equation \eqref{Eq1}), \textit{i.e.,} number of tasks in meta-learning. 
The results in Figure \ref{fig:OOD} (right) showed that 
the MCC metric exceeded 0.85 when 7 tasks appeared in the training data and stabilized when 8 or more tasks appeared. 
This agreed with the theorem that, 
for the pendulum system that had a 3-dimensional parameter vector to be identified in the neural component of the hybrid model with a Gaussian assumption of 2-dimensional sufficient statistics, a minimum of 3*2+1 = 7 distinct “tasks” is needed to identify the system. 

\paragraph{Additional Ablation Analyses of Meta-Hybrid-VAE:}  
In Appendix \ref{sub:ablation1}, 
we provided ablation results on the pendulum system to demonstrate that 
the identifiability results of the presented meta-hybrid-VAE were minimally affected by the number of parameters to be identified, as long as the theoretical condition for identifiability is met.
In Appendix \ref{sub:ablation2}, 
we provided further ablation studies to show that 
the physics-based component did not suffer from the type of un-identifiability discussed in this paper 
as the neural component, 
although the meta-formulation did moderately improve the accuracy of the estimation of the physics-based parameters. 

\begin{figure}[t]
    \centering
    \includegraphics[width=\linewidth]{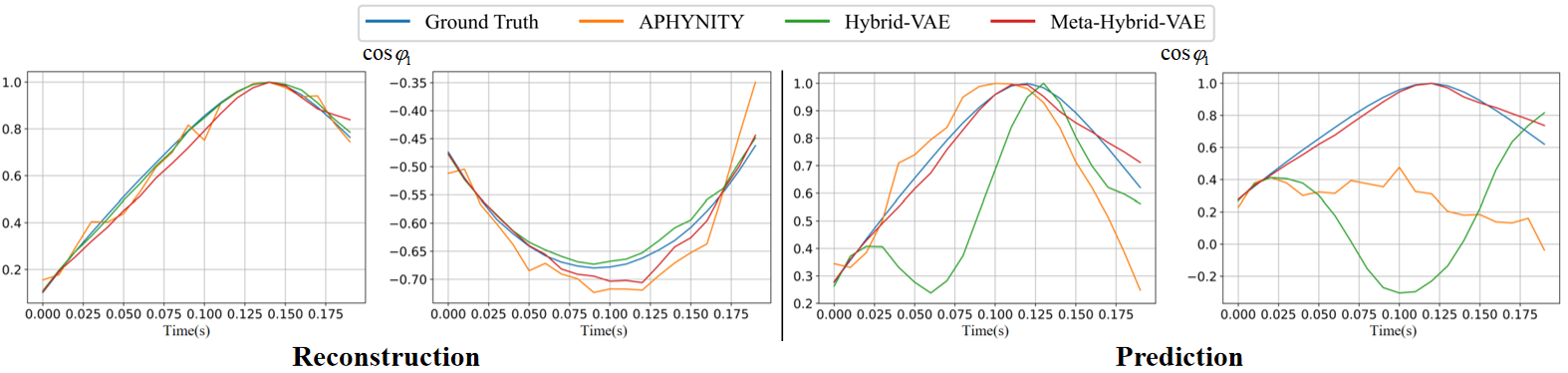}
    \caption{Examples of reconstruction and generation results on real double pendulum data.}
    \label{fig:real}
\end{figure}

\subsection{Real Data of Double Pendulum} 
\label{subsec:real}

\paragraph{Experimental settings:} We used the dataset of a double pendulum introduced by \cite{real-double-pendulum}, which contains 21 videos of the pendulum. Each run lasts approximately 40 seconds and is recorded at 400Hz. We extracted the position of the pendulum limbs from each frame with elementary computer-vision tools.
We divided each video into observations of $\mathbf{x}$'s to consist of 20 temporal frames with a sampling frequency of 100 Hz. Because there is no clear indication of which samples belong to the same data-generation process, we 
use 7 samples preceding the current query sample as the context samples.
We generated 10000, 3400, and 3400 training, validation, and test samples, respectively.

We adopted Equation \eqref{eq:dp} as the physics-component of the hybrid-DGM, 
using the known lengths of the two arms ($L_1=91mm$ and $L_2=70mm$) 
and assuming $\Tilde{m}=1$. 
The total energy of the double pendulum decreases over time in all videos, which is not explained by this physics model. Nor does the physics model consider potential vibrations or errors in extracting the arms’ positions, which 
is expected to be learned by the neural component of the hybrid-DGM with the dimension of $\mathbf{z}_\text{N}$ = 4.

\paragraph{Results:} 
 As ground-truth generative factors are not available in the real data, 
 we relied on prediction MSE as a surrogate for the identifiability metric as informed by results in Sections \ref{subsec:p}-\ref{subsec:dp}. 
While all models delivered comparable reconstruction MSE, 
 meta-hybrid-VAE was able to 
 achieve a prediction MSE at a level similar to reconstruction 
 while the two baselines experienced a deterioration by 2-3 magnitudes. 
 Visual examples are provided in Fig.~\ref{fig:real}. 
This strong performance in real data provides solid evidence for 
the presented meta-formulation to 
establish identifiability of hybrid-DGMs. 

\section{Conclusions \& Discussion}

In this paper, 
we probe the un-identifiability of hybrid-DGMs with unconditioned priors, 
and present meta-learning as a novel solution to construct identifiable hybrid-DGMs 
-- both results were supported by strong theoretical and empirical evidence. Moving forward, this work can be improved as follows. 

Comprehensive ablation studies of the meta-hybrid-VAE are needed to examine the effect of several key hyperparameters, such as the size of $k$-shot context set and the dimension of $\mathbf{z}_\text{N}$. The latter was set to the number of true generative factors in synthetic experiments, and its effect on model performance needs to be further examined. Note that this setting was identical across baselines, thus it had no effect on the significant margins of improvements seen by meta-hybrid-VAE compared to baselines. 
Future ablation studies should also delve into the benefit of hybrid-DGMs in comparison to purely physics-based and purely neural DGMs, 
which we considered out of scope of the current study that focused on establishing the identifiability of hybrid-DGMs both theoretically and empirically.

Regarding the generality of the 
meta-formulation as a solution to identifiable DGMs beyond the hybrid formulation, 
Appendix \ref{app:nonhybrid} presented an initial investigation on the performance of the presented meta-formulation on a general non-hybrid VAE on a synthetic dataset of non-stationary Gaussian time-series, in comparison to the identifiable-VAE \cite{iVAE} constructed using known class labels to condition the generative model. 

Future works could investigate this effect in a broader context. 
Finally, 
while the meta-formulation foregoes explicitly observed auxiliary variable, 
it does assume the ability to pair context and query samples from the same data generation process. 
This is a common assumption in meta-learning and, 
in scenarios where this knowledge is not evident, 
simply pairing by preceding temporal samples was shown to be effective in our real-data experiments. 
In scenarios where no information is available for pairing, the presented meta-solution will not be applicable. 

\begin{ack}
This study is supported by the National Key Research and Development Program of China(No: 2020AAA0109502), the Talent Program of Zhejiang Province (No: 2021R51004), the NIH NHLBI grant R01HL145590, the NIH NINR grant R01NR01830, and NSF grant OAC-2212548.

\end{ack}

\bibliographystyle{unsrt}
\bibliography{Styles/reference}

\newpage
\appendix

\section{Details on Experimental Data Generation}
\label{app:data}

\subsection{Forced Damped Pendulum} 
\label{subsec:p}

 We generated data 
from the \textit{full physics} function in Equation \eqref{Eq1}, 
with observed $\mathbf{x}=[\varphi_0, \dots, \varphi_i,\dots,\varphi_T], \varphi_i=\varphi(i\Delta t)$, $T=49, \Delta t=0.05$. We randomly sampled the initial condition $\varphi_0 \in (-1.57, 1.57]$ (with $\dot{\varphi}_0=0$ fixed). 
Physics parameters $[\omega, \xi, A, \phi]$ to be identified were varied as follows: 
$\omega \in [0.785, 3.14]$ , $\xi \in [0.0, 0.8]$, $A \in [0, 40]$ and $\phi \in [3.14, 6.28]$. We generated total 60000 samples and separated them into a training, validation, and test sets with 4,0000, 10000, and 1,0000 samples, respectively. 
In all hybrid-DGMs, 
we considered the first two terms n in Equation \eqref{Eq1}
as the \textit{partial physics}, thus 
$\mathbf{z}_\text{P} = \omega$ and 
$\mathbf{z}_\text{N}$ with a dimension of 3 is expected to recover 
$[\xi, A, \phi]$.  
This setting induces an additive error in the prior physics. In OOD setting, we set OOD $\mathbf{z}_\text{P}$ as $\omega\in [3.14, 3.5]$  and OOD $\mathbf{z}_\text{N}$ as $\xi \in [0.8, 1.0]$, while keeping the rest of the parameters unchanged.

\subsection{Advection-diffusion System} 

 We generate data from the following \textit{full physics} function:
\begin{equation}
\label{eqn:ad}
    \frac{\partial T}{\partial t}-a\frac{\partial^2T}{\partial s^2}+b\frac{\partial T}{\partial s}=0
\end{equation}
where $s$ is the spatial dimension. The solution $T(s,t)$ was observed 
as %
$\mathbf{x}:=[\mathbf{T}_0, \dots, \mathbf{T}_T]$, where $\mathbf{T}_j:=[T(0, t_j),\dots, T(s_{max}, t_j)]^T$ at $t_j:= j\Delta t$ is uniformly distributed in space with a dimension of 20. 
We set $T=49,\Delta t=0.02$, 
the boundary condition as $T(0,t)=T(s_{max}, t)=0$, 
and the initial condition as $T(s,0)=c\sin(\pi s/s_{max})$ with 
$c\in [0.5, 1.5]$.  
Physics parameters to be identified 
$[a, b]$ were sampled as: 
$a\in [0.01, 0.1]$ and $b\in [0.01, 0.1]$. We generated %
20000, 5000, and 5000 training, validation, and test samples, respectively. 
In all hybrid-DGMs, we set \textit{partial physics} as the first two terms of 
Equation \eqref{eqn:ad}, thus 
$z_p = a$ and $z_\text{N}$ with a dimension of 1 is expected to recover 
$b$. This represents an additive error in the prior physics. 

\subsection{Double Pendulum} 
\label{subsec:dp}

We generated data from the following \textit{full physics} function:
\begin{equation}
\label{eq:dp}
    \frac{d}{dt}\left [ \begin{matrix} \dot{\varphi}_1 \\ \dot{\varphi}_2\end{matrix} \right ] = \left [ \begin{matrix}
    \frac{G\sin\varphi_2\cos(\varphi_1-\varphi_2)-\sin(\varphi_1-\varphi_2)(L_1\dot{\varphi}_1^2\cos(\varphi_1-\varphi_2)+L_2\dot{\varphi}_2^2) - (\Tilde{m}+1)G\sin\varphi_1}{L_1(\Tilde{m}+\sin(\varphi_1-\varphi_2)^2)} \\
    \frac{(\Tilde{m}+1)(L_1\dot{\varphi}_1^2\sin(\varphi_1-\varphi_2)-G\sin\varphi_2+G\sin\varphi_1\cos(\varphi_1-\varphi_2))+L_2\dot{\varphi}_2^2\cos(\varphi_1-\varphi_2)\sin(\varphi_1-\varphi_2)}{L_2(\Tilde{m}+\sin(\varphi_1-\varphi_2)^2)}
    \end{matrix} \right ] 
\end{equation}
where $\Tilde{m}=m_1/m_2$. 
The observed $\mathbf{x}:=[(\varphi_{1,0}, \varphi_{2,0}), 
\dots, 
(\varphi_{1,j}, \varphi_{2,j})
\dots, (\varphi_{1,T},\varphi_{2,T})]$, where $(\varphi_{1,j}, \varphi_{2,j})$ is the value of a solution at $t_j:=j\Delta t$ with $T=79, \Delta t=0.025, L_1=L_2=1.0$. 
We randomly sampled the initial condition $\varphi_{1,0}, \varphi_{2,0} \in (-1.57, 1.57]$ (with $\dot{\varphi}_{1,0}=\dot{\varphi}_{2,0}=0$).  
Physical parameters $[G, \Tilde{m}]$ to be identified were sampled as follows: 
$G \in [5.0, 15.0]$ and $\Tilde{m} \in [0.5, 1.5]$. We generated 20000, 5000, and 5000 training, validation, and test samples, respectively.
In all hybrid-DGMs, we set \textit{partial physics} to be Equation \eqref{eq:dp} with
$\Tilde{m}=1$, 
thus 
$z_p = G$ and $z_\text{N}$ with a dimension of 1 is expected to recover 
$\Tilde{m}$. 
This induces a multiplicative error in the prior physics. 

\section{The proof of Theorem 1}
\paragraph{Theorem 1: }\textit{Assume we observe data sampled from 
$p_\theta(\mathbf{x}, \mathbf{z}_\text{P}, \mathbf{z}_\text{N}|\mathcal{D}^s)$ 
as defined in 
Equations \eqref{Eq:generating}-\eqref{Eq:exp-distribution}
with parameters 
$\phi = (\theta, \mathbf{T}, \pmb{\lambda}_\zeta)$.}
\textit{Assume the following holds:}
\textit{\begin{enumerate}[(i)]
    \item \textit{The set $\left \{ \mathbf{x} \in \mathcal{X} | \varphi_\varepsilon(\mathbf{x})=0 \right \}$ has measure zero, where $\varphi_\varepsilon$ is the characteristic function of the density $p_\varepsilon$ defined in 
$p_\varepsilon(\mathbf{x}-\mathcal{F}_\theta
    [ f_\text{P}, f_{\text{N}_\theta}; \mathbf{z}_\text{P}, \mathbf{z}_\text{N} ])$}.
    \item \textit{The hybrid mixing function $\mathcal{F}_\theta$ is injective.}
    \item \textit{The sufficient function $T_{i,j}$ are differentiable almost everywhere, and linearly independent on any subset of $\mathcal{X}$ of measure greater than zero.}
    \item \textit{There exist $de+1$ distinct context sets $\mathcal{D}^s_0, \mathcal{D}^s_1,\dots, \mathcal{D}^s_{de}$ such that the $de \times de$ matrix $\mathbf{L}$ defined as follows is invertible:
    }
    \begin{equation}
    \mathbf{L} = (\pmb{\lambda}(\mathcal{D}^{s,1})-\pmb{\lambda}(\mathcal{D}^{s,0}), \dots, \pmb{\lambda}(\mathcal{D}^{s,de})-\pmb{\lambda}(\mathcal{D}^{s,0}))
\end{equation}
\end{enumerate}}

\textit{Then the parameters $\phi=(\theta, \mathbf{T}, \pmb{\lambda}_\zeta)$ are $\sim_A$-identifiable.} 

\textbf{Proof: }This proof mainly refers to \cite{iVAE}. For simplicity, we denote $\mathcal{F}_\theta$ as $\mathbf{f}$, $\mathcal{F}_{\Tilde{\theta}}$ as $\Tilde{\mathbf{f}}$, $\pmb{\lambda}_{\zeta}$ as $\pmb{\lambda}$ and $\pmb{\lambda}_{\Tilde{\zeta}}$ as $\Tilde{\pmb{\lambda}}$.
Suppose we have two sets of parameters $(\mathbf{f},\mathbf{T}, \pmb{\lambda})$ and $(\Tilde{\mathbf{f}},\Tilde{\mathbf{T}}, \Tilde{\pmb{\lambda}})$ such that $p_{\mathbf{f},\mathbf{T}, \pmb{\lambda}}(\mathbf{x}|\mathcal{D}^s)=p_{\Tilde{\mathbf{f}},\Tilde{\mathbf{T}}, \Tilde{\pmb{\lambda}}}(\mathbf{x}|\mathcal{D}^s)$ for all pairs $(\mathbf{x}, \mathcal{D}^s)$. Then:
\begin{align}
    \int_{\mathcal{Z}}p_{\mathbf{T}, \pmb{\lambda}}(\mathbf{z}|\mathcal{D}^s)p_{\mathbf{f}}(\mathbf{x}|\mathbf{z})d\mathbf{z}& =\int_{\mathcal{Z}}p_{\Tilde{\mathbf{T}}, \Tilde{\pmb{\lambda}}}(\mathbf{z}|\mathcal{D}^s)p_{\Tilde{\mathbf{f}}}(\mathbf{x}|\mathbf{z})d\mathbf{z} \\
    \int_{\mathcal{Z}}p_{\mathbf{T}, \pmb{\lambda}}(\mathbf{z}|\mathcal{D}^s)p_\varepsilon(\mathbf{x}-\mathbf{f}(\mathbf{z}))d\mathbf{z}&=\int_{\mathcal{Z}}p_{
    \Tilde{\mathbf{T}}, \Tilde{\pmb{\lambda}}}(\mathbf{z}|\mathcal{D}^s)p_\varepsilon(\mathbf{x}-\Tilde{\mathbf{f}}(\mathbf{z}))d\mathbf{z} \\
    \int_\mathcal{X}p_{\mathbf{T}, \pmb{\lambda}}(\mathbf{f}^{-1}(\bar{\mathbf{x}})|\mathcal{D}^s)\text{vol} J_{\mathbf{f}^{-1}}(\bar{\mathbf{x}})p_\varepsilon(\mathbf{x}-\bar{\mathbf{x}}) d\bar{\mathbf{x}} &= \int_\mathcal{X}p_{\Tilde{\mathbf{T}}, \Tilde{\pmb{\lambda}}}(\Tilde{\mathbf{f}}^{-1}(\bar{\mathbf{x}})|\mathcal{D}^s)\text{vol} J_{\Tilde{\mathbf{f}}^{-1}}(\bar{\mathbf{x}})p_\varepsilon(\mathbf{x}-\bar{\mathbf{x}})d\bar{\mathbf{x}} \\
    \int_{\mathbb{R}^d}\Tilde{p}_{\mathbf{T}, \pmb{\lambda}, \mathbf{f}, \mathcal{D}^s}(\bar{\mathbf{x}})p_\varepsilon(\mathbf{x}-\bar{\mathbf{x}})d\bar{\mathbf{x}} &= \int_{\mathbb{R}^d}\Tilde{p}_{\Tilde{\mathbf{T}}, \Tilde{\pmb{\lambda}}, \Tilde{\mathbf{f}}, \mathcal{D}^s}(\bar{\mathbf{x}})p_\varepsilon(\mathbf{x}-\bar{\mathbf{x}})d\bar{\mathbf{x}} \\
    (\Tilde{p}_{\mathbf{T}, \pmb{\lambda}, \mathbf{f}, \mathcal{D}^s}*p_{\varepsilon})(\mathbf{x})&=(\Tilde{p}_{\Tilde{\mathbf{T}}, \Tilde{\pmb{\lambda}}, \Tilde{\mathbf{f}}, \mathcal{D}^s}*p_{\varepsilon})(\mathbf{x}) \\
    F[\Tilde{p}_{\mathbf{T}, \pmb{\lambda}, \mathbf{f}, \mathcal{D}^s}](\omega)\varphi_\varepsilon(\omega)&=F[\Tilde{p}_{\Tilde{\mathbf{T}}, \Tilde{\pmb{\lambda}}, \Tilde{\mathbf{f}}, \mathcal{D}^s}](\omega)\varphi_\varepsilon(\omega) \\
    F[\Tilde{p}_{\mathbf{T}, \pmb{\lambda}, \mathbf{f}, \mathcal{D}^s}](\omega)&=F[\Tilde{p}_{\Tilde{\mathbf{T}}, \Tilde{\pmb{\lambda}}, \Tilde{\mathbf{f}}, \mathcal{D}^s}](\omega) \\
    \label{Eq: conclusion}
    \Tilde{p}_{\mathbf{T}, \pmb{\lambda}, \mathbf{f}, \mathcal{D}^s}(\mathbf{x})&=\Tilde{p}_{\Tilde{\mathbf{T}}, \Tilde{\pmb{\lambda}}, \Tilde{\mathbf{f}}, \mathcal{D}^s}(\mathbf{x})
\end{align}

By taking the logarithm on both sides of equation (\ref{Eq: conclusion}) and replacing $p_{\mathbf{T}, \pmb{\lambda}}$ by its expression from (\ref{Eq:conditioned prior}), we get:
\begin{equation}
\label{Eq: out}
\begin{aligned}
    & \log \text{vol} J_{\mathbf{f}^{-1}}(\mathbf{x})+\sum_{i=1}^n(\log Q_i(f_i^{-1}(\mathbf{x}))-\log Z_i(\mathcal{D}^s))+\sum_{j=1}^k T_{i,j}(f_i^{-1}(\mathbf{x})\lambda_{i,j}(\mathcal{D}^s))= \\
    & \log \text{vol} J_{\Tilde{\mathbf{f}}^{-1}}(\mathbf{x})+\sum_{i=1}^n(\log \Tilde{Q}_i(\Tilde{f}_i^{-1}(\mathbf{x}))-\log \Tilde{Z}_i(\mathcal{D}^s))+\sum_{j=1}^k \Tilde{T}_{i,j}(\Tilde{f}_i^{-1}(\mathbf{x}) \Tilde{\lambda}_{i,j}(\mathcal{D}^s))
\end{aligned}
\end{equation}

Let $\mathcal{D}^s_0, \mathcal{D}^s_1, \dots, \mathcal{D}^s_{de}$ be the context sets provided by assumption (iv) of the Theorem, and define $\bar{\pmb{\lambda}}(\mathcal{D}^s_l)=\pmb{\lambda}(\mathcal{D}^s_l)-\pmb{\lambda}(\mathcal{D}^s_0)$. We plug each of those $\mathcal{D}^s_l$ in (\ref{Eq: out}) to obtain $de+1$ such equations. We subtract the first equation for $\mathcal{D}^s_0$ from the remaining $de$ equations to get for $l=1,\dots, de$:
\begin{equation}
    \label{Eq:001}
    \left \langle \mathbf{T}(\mathbf{f}^{-1}(\mathbf{x})), \bar{\pmb{\lambda}}(\mathcal{D}^s_l)  \right \rangle + \sum_i\log \frac{Z_i(\mathcal{D}^s_0)}{Z_i(\mathcal{D}^s_l)} = 
    \left \langle \Tilde{\mathbf{T}}(\Tilde{\mathbf{f}}^{-1}(\mathbf{x})), \bar{\Tilde{\pmb{\lambda}}}(\mathcal{D}^s_l)  \right \rangle + \sum_i\log \frac{\Tilde{Z}_i(\mathcal{D}^s_0)}{\Tilde{Z}_i(\mathcal{D}^s_l)}
\end{equation}

Let $L$ be the matrix defined in assumption (iv), and $\Tilde{L}$ similarly defined for $\Tilde{\pmb{\lambda}}$. Define $b_l=\sum_i\log\frac{\Tilde{Z}_i(\mathcal{D}^s_0)Z_i(\mathcal{D}^s_l)}{Z_i(\mathcal{D}^s_0)\Tilde{Z}_i(\mathcal{D}^s_l)}$ and $\mathbf{b}$ the vector of all $b_l$ for $l=1, \dots, de$. Expressing equation (\ref{Eq:001}) for all points $\mathbf{u}_l$ in matrix form, we get:
\begin{equation}
    L^T\mathbf{T}(\mathbf{f}^{-1}(\mathbf{x}))=\Tilde{L}^T\Tilde{\mathbf{T}}(\Tilde{\mathbf{f}}^{-1}(\mathbf{x}))+\mathbf{b}
\end{equation}
We multiply both sides by the transpose of the inverse of $L^T$ from the left to find:
\begin{equation}
    \label{eq:goal}
    \mathbf{T}(\mathbf{f}^{-1}(\mathbf{x}))=\mathbf{A}\Tilde{\mathbf{T}}(\Tilde{\mathbf{f}}^{-1}(\mathbf{x}))+\mathbf{c}
\end{equation}
where $\mathbf{A}=L^{-T}\Tilde{L}$ and $\mathbf{c}=L^{-T}\mathbf{b}$.

Now by the definition of $\mathbf{T}$ and according to assumption (iii), its Jacobian exists and is an $de \times d$ matrix of rank $d$. This implies that the Jacobian of $\Tilde{\mathbf{T}}\circ \Tilde{\mathbf{f}}^{-1}$ exists and is of rank $d$ and so is $A$.  We distinguish two cases:\\
(1) If $e=1$, then this means that $A$ is invertible (because $A$ is $d \times d$). \\
(2) If $e>1$, define $\bar{\mathbf{x}} = \mathbf{f}^{-1}(\mathbf{x})$ and $\mathbf{T}_i(\bar{\mathbf{x}}_i) = (\mathbf{T}_{i,1}(\bar{\mathbf{x}}_i), \dots, \mathbf{T}_{i,e}(\bar{\mathbf{x}}_i))$. According to Lemma 3 in \cite{iVAE}, for each $i \in [1, \dots, d]$ there exist $e$ points $\bar{\mathbf{x}}_i^1, \dots, \bar{\mathbf{x}}_i^e$ such that $(\mathbf{T}'_i(\bar{\mathbf{x}}_i^1), \dots, \mathbf{T}'_i(\bar{\mathbf{x}}_i^e))$ are linearly independent. Collect those points into $e$ vectors $(\bar{\mathbf{x}}^1, \dots, \bar{\mathbf{x}}^e)$, and concatenate the $e$ Jacobians $\mathbf{J}_{\mathbf{T}}(\bar{\mathbf{x}}^l)$ evaluated at each of those vectors horizontally into the matrix $Q = (\mathbf{J}_{\mathbf{T}}(\bar{\mathbf{x}}^1), \dots, \mathbf{J}_{\mathbf{T}}(\bar{\mathbf{x}}^e))$ (and similarly define $\tilde{Q}$ as the concatenation of the Jacobians of $\tilde{\mathbf{T}}(\mathbf{f}^{-1}(\tilde{\mathbf{f}}(\bar{\mathbf{x}})))$ evaluated at those points). Then the matrix $Q$ is invertible (through a combination of Lemma 3 and the fact that each component of $\tilde{\mathbf{T}}$ is univariate). By differentiating equation (\ref{eq:goal}) for each $\mathbf{x}^l$, we get (in matrix form): $Q = \mathbf{A} \tilde{Q}$. The invertibility of $Q$ implies the invertibility of $\mathbf{A}$ and $\tilde{Q}$.

Hence, equation (\ref{eq:goal}) and the invertibility of $\mathbf{A}$ mean that $(\mathbf{f}, \mathbf{T}, \pmb{\lambda}) \sim_A (\tilde{\mathbf{f}}, \tilde{\mathbf{T}}, \tilde{\pmb{\lambda}})$

\section{Ablation Analyses of Meta-Hybrid-VAE}

\subsection{Effect of the Dimension of Unknown Parameters}
\label{sub:ablation1}
Here, we provided ablation results on the pendulum system to demonstrate that the identifiability results of the presented meta-hybrid-VAE were minimally affected by the number of parameters to be identified, as long as the theoretical condition for identifiability is met. The results can be seen in Table \ref{tb:ablation1}.

\begin{table}[]
\centering
\caption{The results of the different dimensions of unknown parameters}
\label{tb:ablation1}
\begin{tabular}{|c|c|c|c|c|}
\hline
\begin{tabular}[c]{@{}c@{}}The number of \\ parameters\end{tabular} 

& MSE of $\mathbf{z}_\text{P} \downarrow$ & MCC $\uparrow$
& MSE of $\mathbf{x} $ (Rec) & MSE of $\mathbf{x} $ (Pre) \\ \hline
2 & 1.22(0.08)e-2 & 0.97(0.01) & 1.22(0.08)e-2 & 2.02(0.13)e-2 \\ \hline
3 & 3.62(0.01)e-2 & 0.99(0.00) & 1.29(0.05)e-2 & 5.01(0.00)e-2 \\ \hline
4 & 1.59(0.07)e-2 & 0.99(0.00) & 2.91(0.06)e-2 & 6.85(1.11)e-2    \\ \hline
\end{tabular}
\end{table}

\subsection{Meta-component on cp}
\label{sub:ablation2}
Here, we provided further ablation studies to show that 
the physics-based component did not suffer from the type of un-identifiability discussed in this paper 
as the neural component, 
although the meta-formulation did moderately improve the accuracy of the estimation of the physics-based parameters. The results can be seen in Table \ref{tb:ablation2}.
\begin{table}[]
\centering
\caption{Comparison results of Non-meta and Meta}
\label{tb:ablation2}
\begin{tabular}{|c|c|c|c|c|}
\hline
Method
& MSE of $\mathbf{z}_\text{P} \downarrow$ & MCC $\uparrow$
& MSE of $\mathbf{x} $ (Rec) & MSE of $\mathbf{x} $ (Pre) \\ \hline
Non-Meta $\mathbf{z}_\text{P}$ & 9.17(1.33)e-3 & 0.99(0.00) & 3.28(0.17)e-2 & 8.82(2.03)e-2 \\ \hline
Meta $\mathbf{z}_\text{P}$ & 5.63(0.42)e-3 & 0.99(0.00) & 3.18(0.19)e-2 & 3.37(0.00)e-2 \\ \hline
\end{tabular}
\end{table}

\section{Results on General Non-Hybrid VAEs}
\label{app:nonhybrid}
We compare our meta-formulation of the non-hybrid VAE (Meta-VAE) compared to the original VAE and identifiable VAE (iVAE) on a synthetic datasets about non-stationary Gaussian time-series in \cite{iVAE}. To evaluate the performance of the method, we compute the MCC between the original sources and the corresponding latent samples from the learned posterior. The results can be seen in Table \ref{tb:general}.

\begin{table}[h]
\caption{The results on General Non-Hybrid VAEs}
\label{tb:general}
\centering
\begin{tabular}{|c|c|c|c|}
\hline
    & VAE & iVAE & Meta-VAE \\ \hline
MCC &  0.67(0.04)   &  0.91(0.03)    &  0.88(0.00)        \\ \hline
\end{tabular}
\end{table}

\section{Strong MCC results}
\label{app:strongMCC}
We also evaluate proposed model on a more strong identifiable relation $\sim_P$ defined as follow: 
\paragraph{Definition 3: }\textit{Let $\sim$ be the equivalence relation on $\Phi$ defined the same as \ref{eq:identifiable_A}: }
\begin{equation}
    (\theta,\mathbf{T}, \pmb{\lambda}_\zeta) \sim (\Tilde{\theta}, \Tilde{\mathbf{T}}, \pmb{\lambda}_{\Tilde{\zeta}}) \Leftrightarrow \exists \mathbf{A},\mathbf{c}: \mathbf{T}(\mathcal{F}_{\theta}^{-1}(\mathbf{x}))=\mathbf{A}\Tilde{\mathbf{T}}(\mathcal{F}_{\Tilde{\theta}}^{-1}(\mathbf{x}))+\mathbf{c}, \forall\mathbf{x} \in \mathcal{X}
\end{equation}
\textit{If $\mathbf{A}$ is a \textit{block} permutation matrix, we denote this relation by $\sim_P$.}

The theory according to $\sim_P$ is following:
\paragraph{Theorem 2: }\textit{($e\ge2$) Assume the hypotheses of Theorem 1 hold and that $e\ge2$. Further assume:}
\textit{\begin{enumerate}[(2.i)]
    \item The sufficient statistics $T_{i,j}$ in Equation \ref{Eq:exp-distribution} are twice differentiable.
    \item The mixing function $\mathcal{F}_\theta$ has all second order cross derivatives.
\end{enumerate}
then the parameters $(\theta,\mathbf{T}, \pmb{\lambda}_\zeta)$ are $\sim_P$-identifiable.}

\paragraph{Theorem 3: }\textit{$(e=1)$} Assume the hypotheses of Theorem 1 hold and that $e=1$. Further assume:
\textit{\begin{enumerate}[(3.i)]
    \item The sufficient statistics $T_{i,1}$ not monotonic.
    \item All partial derivatives of $\mathcal{F}_\theta$ are continuous.
\end{enumerate}
then the parameters $(\theta,\mathbf{T}, \pmb{\lambda}_\zeta)$ are $\sim_P$-identifiable.}\\

The proof of Theorem 2 and 3 can be found in \cite{iVAE}. Then to evaluate the performance on $\sim_P$, we calculate the strong MCC between learned latent variables and ground truth variables (Strong MCC Ground Truth) and strong MCC between runs of the model (Strong MCC Different Runs) \cite{iVAE, GCL, ICE-BeeM}. The results can be found in Table \ref{tb:strongMCC} and we provide a visualization example of forced damped pendulum system to show the relation of learned latent variables between two different runs.

\begin{figure}[t]
    \centering
    \includegraphics[width=0.75\linewidth]{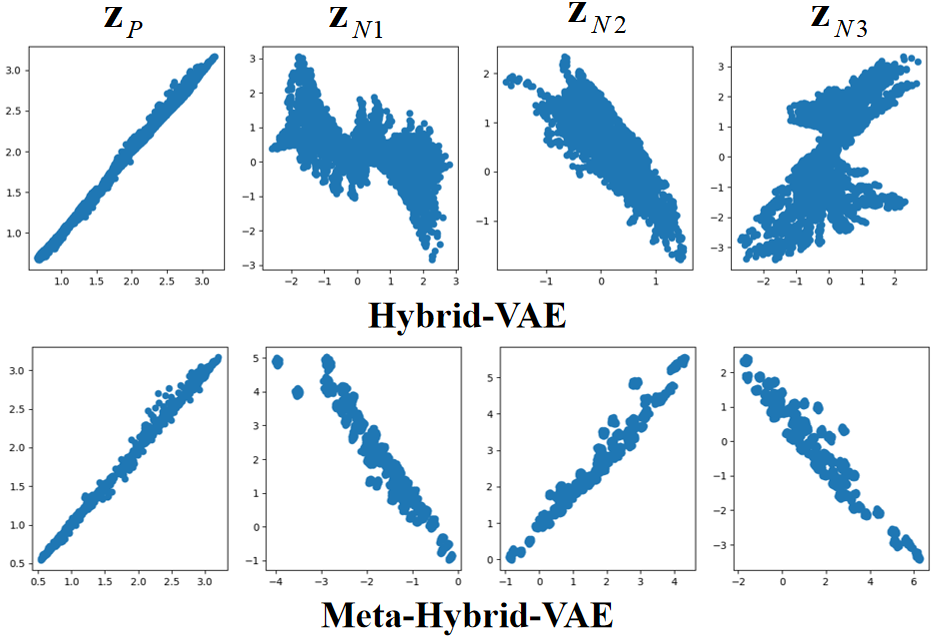}
    \caption{Visual samples of learned latent variables between different runs.}
    \label{fig:identifiable-P}
\end{figure}

\begin{table}[t]
\caption{The results of strong MCC}
\label{tb:strongMCC}
\resizebox{\linewidth}{!}{
\begin{tabular}{|c|cc|cc|cc|}
\hline
\multirow{2}{*}{} & \multicolumn{2}{c|}{Forced Damped Pendulum}  & \multicolumn{2}{c|}{Advection-Diffusion System}  & \multicolumn{2}{c|}{Double Pendulum} \\ \cline{2-7} 
& \multicolumn{1}{c|}{\begin{tabular}[c]{@{}c@{}}Strong MCC\\ Ground Truth\end{tabular}} & \begin{tabular}[c]{@{}c@{}}Strong MCC\\ Different Runs\end{tabular} & \multicolumn{1}{c|}{\begin{tabular}[c]{@{}c@{}}Strong MCC\\ Ground Truth\end{tabular}} & \begin{tabular}[c]{@{}c@{}}Strong MCC\\ Different Runs\end{tabular} & \multicolumn{1}{c|}{\begin{tabular}[c]{@{}c@{}}Strong MCC\\ Ground Truth\end{tabular}} & \begin{tabular}[c]{@{}c@{}}Strong MCC\\ Different Runs\end{tabular} \\ \hline
APHYNITY  & \multicolumn{1}{c|}{0.38(0.02)}  & 0.81(0.08)  & \multicolumn{1}{c|}{0.86(0.05)}  &  0.83(0.04) & \multicolumn{1}{c|}{0.50(0.00)}  & 0.73(0.00) \\ \hline
Hybrid-VAE  & \multicolumn{1}{c|}{0.29(0.00)}  & 0.80(0.01) & \multicolumn{1}{c|}{0.80(0.12)}   & 0.79(0.11)  & \multicolumn{1}{c|}{0.50(0.00)} & 0.72(0.00)                                                                     
\\ \hline
Meta-Hybrid-VAE   & \multicolumn{1}{c|}{\textbf{0.55(0.02)}}  &  \textbf{0.90(0.06)}  & \multicolumn{1}{c|}{\textbf{0.97(0.03)}}   &  \textbf{0.96(0.00)}  & \multicolumn{1}{c|}{\textbf{0.96(0.00)}} & \textbf{1.00(0.00)}            \\ \hline
\end{tabular}}
\end{table}

\end{document}